\documentclass{article}
\usepackage{spconf,amsfonts,graphicx}

\newcommand{\RomanNumeralCaps}[1]
    {\MakeUppercase{\romannumeral #1}}
\usepackage{subfigure}
\usepackage{amsmath}
\usepackage{multirow}

\title{Spectral~Oversubtraction?~An~Approach~for~Speech~Enhancement~after~Robot~Ego~Speech~Filtering in Semi-Real-Time
}
\twoauthors
 {Yue Li\sthanks{Thanks to Chinese Scholarship Council for funding.}, Koen V. Hindriks}
	{Vrije Universiteit Amsterdam\\
	Social AI\\
	De Boelelaan 1105 Amsterdam.}
 {Florian A. Kunneman}
	{Utrecht Universiteit\\
	Languages, Literature and Communication\\
	Heidelberglaan 8 Utrecht.}
\begin{document}
%\ninept
%
\maketitle
\begin{abstract}
    
Spectral subtraction, widely used for its simplicity, has been employed to address the Robot Ego Speech Filtering (RESF) problem for detecting speech contents of human interruption from robot's single-channel microphone recordings when it is speaking. However, this approach suffers from oversubtraction in the fundamental frequency range (FFR), leading to degraded speech content recognition.
To address this, we propose a Two-Mask Conformer-based Metric Generative Adversarial Network~(CMGAN) to enhance the detected speech and improve recognition results. Our model compensates for oversubtracted FFR values with high-frequency information and long-term features and then de-noises the new spectrogram. In addition, we introduce an incremental processing method that allows semi-real-time audio processing with streaming input on a network trained on long fixed-length input. Evaluations of two datasets, including one with unseen noise, demonstrate significant improvements in recognition accuracy and the effectiveness of the proposed two-mask approach and incremental processing, enhancing the robustness of the proposed RESF pipeline in real-world HRI scenarios.
    % what did we do?

    % why did we do it?

    % how did we do it?

    % What did we find?

    % why is this important?

\end{abstract}

\section{Introduction}
\label{section:1}
% 1st way to start 
% During a conversational Human-Robot Interaction~(HRI), humanoid robots, as sophisticated as Pepper, cannot interpret any human speech interrupting, even though they are equipped with the most state-of-the-art Automatic Speech Recognition~(ASR) systems~\cite{Li_2024}. In order to solve this, a previous paper~\cite{Li_2024} adopted spectral subtraction to filter robot ego speech and estimate overlapping human speech from the spectrogram of the received mixture sound. The effectiveness of the proposed pipeline was demonstrated by the implementation in Pepper in a pilot experiment. 

% 2nd way to start
Spectral subtraction~(SS) is one of the most prevalent signal processing methods for speech enhancement~(SE), comprising the subtraction of the estimated noise spectrum from the recorded signal spectrum~\cite{paliwal2010single}. Due to its simplicity and effectiveness, SS has been widely adopted in various practical applications, such as mobile communication, hearing aids, and speech recognition systems~\cite{das2021fundamentals}. In this study, we focus on the case of %It was also applied to filter 
robot ego speech filtering~(RESF) during human-robot interaction~(HRI) based on a well-known social robot, which is most often used in HRI, Pepper~\cite{Li_2024}. We intend to improve the performance in detecting and recognizing speech contents when a human interrupts robot speech from its embedded single-channel recordings. %make sense of a human's the interruption of a human. %listening and interpreting human's interruption when itself is speaking. 
Such a capability, when done right, can make HRI much more fluent, as the human can barge in as it would do in human-human interactions and would not have to wait until the robot starts listening with the microphone after finishing its utterance. 

% Continue to introduce the problem
%However, d
Despite its widespread use, SS has notable drawbacks, in particular spectral oversubtraction caused by excessive noise spectrum estimation~\cite{upadhyay2015speech}. In dynamic and non-stationary noise environments, accurately estimating the noise spectrum is inherently challenging, leading to distorted enhanced speech~\cite{kleinschmidt2010robust}. The primary issue is %For example, when filtering robot ego speech, SS can result in distorted speech estimation due to 
oversubtraction in the fundamental frequency range~(FFR) when spectrally subtracting Pepper ego speech from its single-channel recordings ~\cite{li2024near}, further leading to misrecognition of words with nasal or plosive sounds by state-of-the-art~(SOTA) automatic speech recognition (ASR) systems~\cite{parikh2005influence}. % This misrecognition hampers the robot's ability to interpret human intentions and limits the application of RESF pipeline~\cite{li2024near} in real-world %the wild 
%HRI scenarios.
% Introduce the gap between traditional research and the topic we focus.
To address this, researchers have developed adaptive noise estimation filters and spectral floor adjustments~\cite{chaudhari2015review}.  Kamath and Loizou~\cite{kamath2002multi} proposed a multi-band non-linear spectral subtraction method,
% ~(MBSS)
%, %in which they 
using a frequency-dependent subtraction factor to account for different types of noise% and made the subtraction process nonlinear over the frequency range
%. %To address variable noise levels and characteristics, Virag~\cite{virag1999single} investigated the use of perceptual properties of the human auditory system to improve the intelligibility and quality of the enhanced speech signal
. Such methods perform well in stationary noise environments or when the signal-to-noise ratio~(SNR) is not less than 10~dB~\cite{upadhyay2015speech}. However, they struggle in non-stationary noise conditions where the SNR is significantly lower, a condition that is quite common to HRI where constraints of humanoid robot design can cause SNR values as low as -20~dB~\cite{li2024near}.

% The requirements for this target work.
To correctly interpret human interruptions during HRI, an SE method is needed that improves the intelligibility and the recognition result of detected speech. Through sophisticated machine learning algorithms, 
SE systems have advanced significantly, manifested in approaches such as DCCRN~\cite{hu2020dccrn} and FullSubNet~\cite{hao2021fullsubnet}. Among these advanced networks, generative adversarial networks~(GAN) come with the advantages of speed - its generator can learn to produce accurate representations in a short time - and conservation - its discriminator maintains most of the information from the original data distribution~\cite{wali2022generative}. 
In 2017, Pascual et al.~\cite{pascual2017segan} proposed SEGAN, one of the initial efforts to apply GAN to speech enhancement. A new avenue for GAN-based SE research was then initiated by Fu et al.~\cite{fu2019metricgan} who proposed MetricGAN to directly optimize the evaluation metrics for this task. %,  proposed MetricGAN, in which an evaluation metric based on human auditory perception was incorporated into the discriminator loss function. This innovation created a new avenue for GAN-based SE research. 
Based on these advancements, Cao et al.~\cite{cao2022cmgan} proposed a Conformer-based MetricGAN~(CMGAN), achieving the highest Perceptual Evaluation of Speech Quality~(PESQ) score in the \textit{VoiceBank+DEMAND} dataset~\cite{thiemann2013diverse} as of 2022. %Its subsequent ablation study provided additional insight in 2024~\cite{abdulatif2024cmgan}. 
% Introduce GAN for speech enhancement and its limitation in the task we are facing
Despite these advancements, the ASR results in enhanced speech have declined rather than improved~\cite{ristea2024icassp}. For example, %Iwamoto et al.~\cite{iwamoto2022bad} investigated the influence of SE-induced artifacts on the final ASR result and proposed reintroducing noise back into enhanced speech to alleviate the influence. 
Donahue et al.~\cite{Donahue2018} proposed FSEGAN and jointly trained it with an ASR system,  %However, in terms of improving the ASR results, they
finding that the usefulness of GANs was limited compared to replacement of the training object. 

To the best of our knowledge, no SE system specifically targets recovery from spectral oversubtraction distortion in the FFR and performs on recorded streaming audio buffers to achieve semi-real-time processing. Neither is there a dataset consisting of speech that is distorted by spectral oversubtraction. Previous work % In addition, the success of translating some words in distorted speech by SOTA ASR systems indicates that the values in the high frequency range~(HFR) are sufficient but not enough to recognize words~\cite{li2024near}. Furthermore, it 
suggests that values in the HFR can be learned to restore the oversubtracted values in the FFR and improve the vulnerability of detected speech to noise.
%Lastly, 
Furthermore, the success of MetricGAN, which was outside the scope of the study by Donahue et al.~\cite{Donahue2018}, shows the potential of GAN-based SE systems to improve the ASR result after SE.

The aim of this work is two-fold: 1. Develop and verify a MetricGAN-based network that can learn from the information in HFR to restore the oversubtracted values in FFR and to improve the ASR results;  2. Improve the robustness of the RESF system that is vulnerable to common ambient noise in real world HRI~\cite{li2024near}. We propose a Two-Mask CMGAN inspired by~\cite{cao2022cmgan}.
% on the basis of previous research~\cite{li2024near}. 
We examine the impact of different inputs for the discriminator network on the ASR results and propose a method that allows a network trained on fixed-length audio segments to process short audio buffers while accessing long-term information. Evaluation on offline datasets with and without additional %real-life 
noise demonstrates the effectiveness and robustness of the proposed pipeline.%, which further strengthens the application reliability of the RESF system in the wild.

\section{Methods}
\label{Section:methods}
\subsection{Two-Mask CMGAN}
% In the  RESF pipeline taken from~\cite{li2024near}, robot speech is spectrally subtracted from the recorded audio spectrogram to estimate human speech interruption. However, because of the significant power difference between the speech of robots and humans, the estimation is distorted because of the oversubtraction in the FFR.

\begin{figure}[t]
    \centering
    \includegraphics[width=0.4\textwidth]{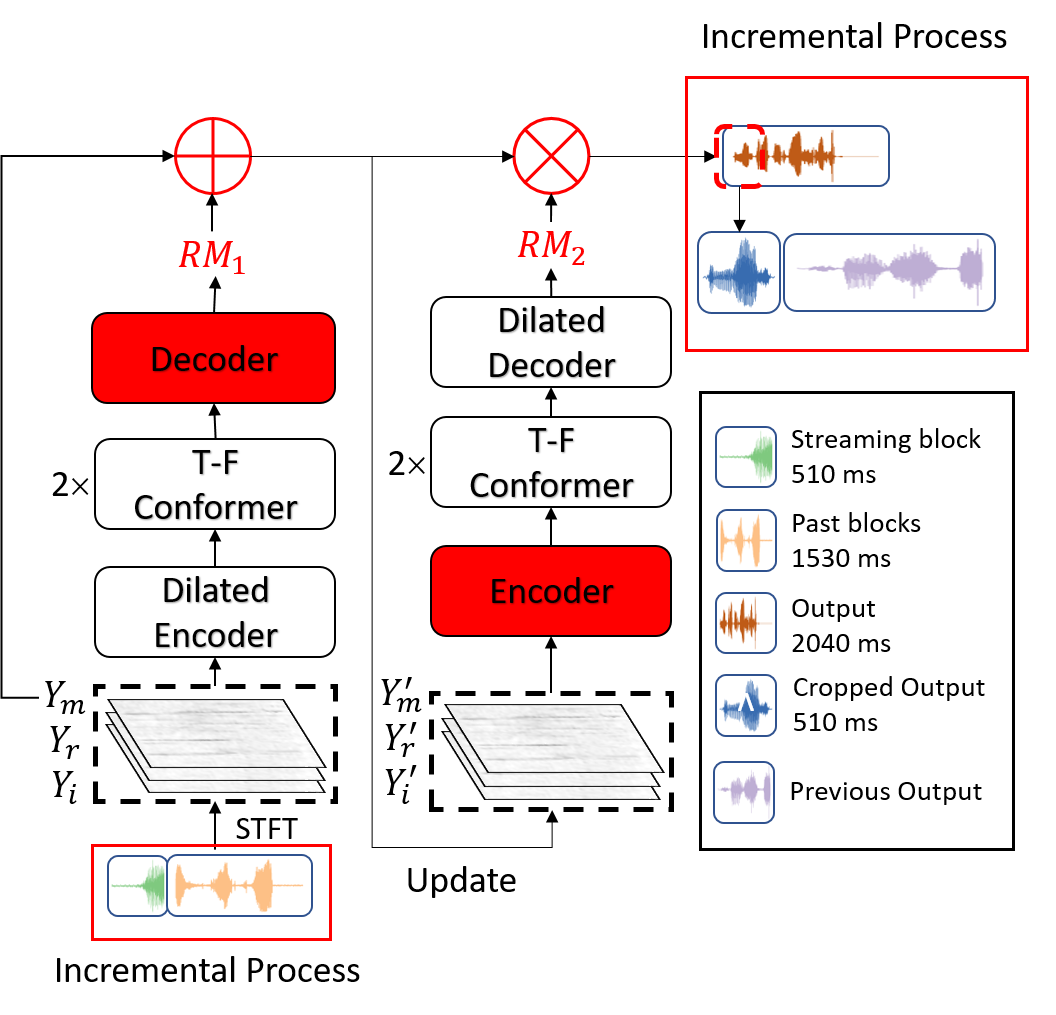}
    \caption{Comparison of the original CMGAN and proposed Two-Mask CMGAN. Components marked in red are proposed in this work and absent in the original CMGAN.}
    \label{fig:2}
\end{figure}

Given that the spectrum of human speech is short-time invariant and the restored information in the higher frequency range is the formant of the oversubtracted information in FFR~\cite{titze1998principles}, we propose a variant of CMGAN: Two-Mask CMGAN to improve the recognition result of detected human interruption speech. The overview of the proposed two-mask CMGAN is shown in Fig.~\ref{fig:2}. 
% In the following, we give a detailed description of its components. 

% \subsubsection{Generator Architecture}

The generator contains three parts: \textit{Encoder}, \textit{Conformer}, and \textit{Decoder}. To verify the hypothesis that oversubtracted values in FFR can be restored based on HFR information, we propose to split the original one ideal ratio mask~(\textit{IRM}) generation procedure into two subprocedures, %Compared to the original CMGAN generator, it can be 
explained in Eq.~\ref{eq:4}.
\begin{equation}
    \hat{S}(tf) = \left(Y(tf) + IRM_1 \right) \cdot IRM_2
\label{eq:4}
\end{equation}
where $IRM_1$ is the compensation ratio mask in the FFR generated according to the information in the HFR, and $IRM_2$ is the denoising ratio mask generated according to the compensated information in the whole frequency range. Because oversubtraction can reduce some values to $0$, which cannot be restored by multiplying the ratio mask, we choose to first generate a compensation ratio mask and add it to the noisy input. Instead of using the \textit{Complex Decoder} as Cao et al. in the original CMGAN \cite{cao2022cmgan}, we adopt Eq.~\ref{eq:6} to update the values in the real and imaginary spectrograms, $Y_r$ and $Y_i$, to reduce the complexity of the network.
\begin{equation}
    \begin{cases} 
    Y^\prime_r = Y^\prime_m \cos{Y_p} \\
    Y^\prime_i = Y^\prime_m \sin{Y_p}
    \end{cases}
\label{eq:6}
\end{equation}

In order to fairly compare the performance of the proposed network with the original CMGAN and alleviate the influence of the number of network parameters, we choose two \textit{Conformer} modules for each mask generation, which in total is the same as the total number of \textit{Conformer} blocks in the original CMGAN.

% \subsubsection{Mel Metric Discriminator}
Due to the limited usefulness of GAN when replacing the training object~\cite{Donahue2018} and the better performance of complex spectrum masking than that of magnitude spectrum masking~\cite{abdulatif2024cmgan}, it is not the best practice to use the same magnitude spectrum as input for the discriminator or directly replace the training object from the complex spectrum to the magnitude spectrum. Therefore, we choose to use the same discriminator architecture as the original CMGAN. However, in order to improve the ASR performance in enhanced speech, we make a trade-off and replace the input of the enhanced speech spectrogram with the input of the SOTA ASR system, \textit{Whisper}\footnote{https://huggingface.co/openai/whisper-large-v3}, to obtain better predictions of the values in FFR. Hence, the discriminator in the proposed architecture aims to mimic the difference between the mel spectrograms of the enhanced and clean speech and further uses it as a part of the loss function.

% \subsubsection{Loss Function}
%Regarding the design of the loss function, 
For loss function design, we adopt the same strategy as Cao et al.~\cite{cao2022cmgan}, which consists of a linear combination of the loss in the TF domain $\mathcal{L}_{\text{TF}}$, the loss in the generator $\mathcal{L}_{\text{GAN}}$ and the resultant waveform loss $\mathcal{L}_{\text{Time}}$. 
% as follows: 
% \begin{equation}
%     \mathcal{L}_{\text{G}} = \gamma_1 \mathcal{L}_{\text{TF}} + \gamma_2 \mathcal{L}_{\text{GAN}}+ \gamma_3 \mathcal{L}_{\text{Time}}
% \label{eq:5}
% \end{equation}
% where $\gamma_1$, $\gamma_2$, and $\gamma_3$ are the corresponding weights to reflect the equal importance between losses.
Based on the work of Mao et al.~\cite{Mao_2017_ICCV}, the adversarial training follows a min-min optimization task over discriminator loss $\mathcal{L}_{\text{D}}$ and generator loss $\mathcal{L}_{\text{GAN}}$. %Unlike 
While Cao et al.~\cite{cao2022cmgan} use the magnitude spectrogram as discriminator input, we opt for the % is replaced from the magnitude spectrogram by the 
mel spectrogram. %Therefore, 
The discriminator loss is as follows:
\begin{equation}
    \begin{split}
        &\mathcal{L}_{\text{D}} = \mathbb{E}_{X_{\text{mel}},\hat{X}_{\text{mel}}}[\parallel D(X_{\text{mel}}, \hat{X}_{\text{mel}}) - 1 \parallel ^2 ]\\
        &\quad + \mathbb{E}_{X_{\text{mel}}, \hat{X}_{\text{mel}}} [ \parallel D(X_{\text{mel}}, \hat{X}_{\text{mel}}) - Q_{\text{PESQ}} \parallel^2  ]
    \end{split}
\end{equation}
where $D$ is the discriminator, $Q_{\text{PESQ}}$ refers to the normalized PESQ score in the range [0,1], $X_{\text{mel}}$ and $\hat{X}_{\text{mel}}$ are the mel-spectrogram of the enhanced and target speech signal with 128 bins.

\vspace{-.2cm}
\subsection{Semi-real-time Incremental Processing}
\label{Section:Semi-real-time}
The attention mechanism was first introduced by Bahdanau et al.~\cite{bahdanau2014neural} to learn the weights between the hidden spaces of different components in a long input sequence.  Its variants %, \textit{Transformer}~\cite{vaswani2017attention}, \textit{Conformer}~\cite{gulati2020conformer}, \textit{etc.}, 
demonstrate its ability to improve the performance of the encoder-decoder model in machine translation. Kim et al.~\cite{kim2020t} and Cao et al.~\cite{cao2022cmgan} respectively implemented \textit{Transformer} and \textit{Conformer} in speech enhancement, and achieved satisfactory results. However, the performance of these models in real-world HRI applications is not satisfying, due to %. One of the reasons is 
the conflict between the requirement for real-time processing during HRI and the fixed length of input during training. %To be specific, 
These speech enhancement models %adopting the attention mechanism 
are trained on fixed length audio input to learn the weights between the hidden spaces, %in this audio. 
which should be long enough to carry the right information. %This input cannot be too short or the weights to be learned are not enough. 
For example, Cao et al. used %the audio with 
fragments of 2 seconds as input during training~\cite{cao2022cmgan}, which is impractical during HRI. %Therefore, based on the physical settings of the popular humanoid robot Pepper, we propose an 
We propose to address this issue by means of %input 
Incremental Processing~(IP). %method (see Fig. % for the model trained on fixed input length to achieve comparable performance. This method is illustrated in Fig.
%~\ref{fig:2}).

From the Pepper NAOqi SDK\footnote{http://doc.aldebaran.com/2-5/naoqi/audio/alaudiodevice.html}, every 170~ms audio buffer recorded by its single channel microphone can be used for local processing. %sent to our model.%a local processing module. 
We adopt 3 buffers as one block, adding up to 510~ms audio, and concatenate it with the blocks previously recorded by Pepper to create an input with a length of 2,040~ms, %which is the same 
similar to the input length reported by %during training as 
Cao et al.~\cite{cao2022cmgan}. This is done with a sliding window, where the earliest block is removed upon each added block, maintaining the fixed input length. %Once obtained, the last 510~ms output is trimmed and concatenated with the blocks from previous ones. 
The first blocks that do not have enough previous recorded audio to add up to 2,040~ms %, we choose 
are padded with 0-values to meet this length. %To demonstrate the effectiveness of the proposed input IP method, we also directly fed the streaming blocks into the network without any processing. 

\section{Experimental Result and Discussions}
\subsection{Data \& Experimental Setup}
We generated two data sets to evaluate the proposed Two-Mask CMGAN and IP procedure.
\vspace{-.2cm}
\subsubsection{Training set and Evaluation Set I}
\label{Section:Dataset}
As discussed in Section~\ref{section:1}, we are not aware of any %to the best of our knowledge, there is no 
public dataset consisting of distorted human speech, of which spectral values in FFR are oversubtracted during speech separation or enhancement. Driven by the goal of enhancing the application of the RESF system, we followed the same dataset generation scheme as~\cite{Li_2024} and created 10,000 triplets using the public datasets Robot Voice\footnote{https://osf.io/v4y6h/} and Librispeech\footnote{https://www.openslr.org/12}. The distorted detection of human speech from the RESF pipeline was considered as %the 
\textit{Distortion Speech} and its corresponding clean human speech was taken as the \textit{Target Speech}. We randomly adopted 1,000 combinations of \textit{Distortion Speech} and \textit{Target Speech} as the \textit{Evaluation Set \RomanNumeralCaps{1}} and the %rest 
other 9,000 as the training set. 

% consisting of three 170~ms audio buffer\footnote{According to the Pepper's manuscript, every 170~ms audio buffer recorded by its microphones can be sent to local to process.}
\subsubsection{Evaluation Set II}
To assess the robustness of the combination of RESF and Two-Mask CMGAN in HRI scenarios where a robot is typically placed in a location with sounds in the background, %, where human interruption speech is not only overlapped with robot speech and its fan noise, but also polluted by various ambient noise, 
we added noise fragments from the Microsoft Scalable Noisy Speech Dataset~(MS-SNSD)~\cite{reddy2019scalable}. % and Robot Voice for further evaluation. 
Specifically, we mixed 61 clean human speech audio files with 17 types of challenging nonstationary noise from MS-SNSD. These can be categorized into six groups: \textit{1) Airport} - AirportAnouncement;
\textit{2) Babble Speech} - Babble, NeighborSpeaking;
\textit{3) Noisy Indoor} - Cafe, Cafeteria, Restaurant, LivingRoom;
\textit{4) Tranquil Indoor} - Copier, Kitchen, Office, Hallway, Typing;
\textit{5) Outdoor} - Park, Square, Station, Traffic;
\textit{6) Vaccuum cleaner} - VacuumCleaner.
Given the provided toolkit\footnote{https://github.com/microsoft/MS-SNSD}, all noisy speech clips were scaled to have nine global SNRs of ${40,35,30,25,20,15,10,5,0}$~dB each, which gave out a 5~hour \textit{Evaluation Set \RomanNumeralCaps{2}}. Then we created noisy overlapping human-robot speech following the same generation scheme as in Section~\ref{Section:Dataset}. 
Subsequently, these noisy overlapping speech segments were cut into 510~ms blocks and fed into the RESF pipeline to obtain the detected human interruption speech blocks. These blocks were finally concatenated as the \textit{Distortion Speech} in \textit{Evaluation Set \RomanNumeralCaps{2}}. Its corresponding clean speech was taken as the \textit{Target Speech}.
% Finally, enhanced estimation of human interruption speech was obtained by Two-Mask CMGAN with these concatenated estimation blocks under the proposed IP method. We must emphasize that all of these added noises are absent from the training set. 

\subsubsection{Experimental Set-up}

Using the training set and the two evaluation sets, we compared the performance of our proposed Two-Mask CMGAN approach to that of the pretrained CMGAN\footnote{https://github.com/ruizhecao96/CMGAN} and the CMGAN retrained on our training set. We set the ASR result on the detected human speech from RESF as the baseline to compare which method would improve the ASR result. In addition, we tested how well both approaches performed without IP, and how well the two-mask CMGAN performed without the mel discriminator. During training, we randomly cut the audio files to 2,040~ms to align with the same input length as~\cite{cao2022cmgan}. To evaluate our proposed IP
% in semi-real-time
, we cut the input into 510~ms blocks and fed them to the models as described in Section~\ref{Section:Semi-real-time}. The performance of the models in improving the interpretation of the detected human interruption speech is evaluated by calculating the WER after applying ASR to the enhanced sound fragment. We adopted Whisper as the ASR system to translate all detected speech by RESF, enhanced speech by CMGAN and Two-Mask CMGAN, and target human speech. The recognition results of the \textit{Target Speech} were taken as ground truth to alleviate the influence of the choice of the ASR system. %Table \ref{tab:1} shows 
We report on the mean WER and standard deviations, as well as the percentage of files whose WER is lower than 20\%, which indicates that no more than two words were misrecognized, inserted, or replaced in a 10-word utterance. We selected this because we found that only a single word was misidentified as a combination of two words in most cases.

%We report on the mean and standard deviation~(SD) values of the WER. We also report the percentage of files whose WER is lower than 20~\%, which indicates that no more than two words were misrecognized, inserted, or replaced in a ten-word utterance.

%To evaluate the performance in improving the interpretation of the estimated human interruption speech, we used WER as the evaluation metric and compared the performance of the proposed method and that of the pretrained CMGAN\footnote{https://github.com/ruizhecao96/CMGAN}. %To further compare the effectiveness of the two-mask generation method in enhancing spectral oversubtracted speech, 
%We also retrained the original CMGAN on the same training set. In addition, we included the Two-Mask CMGAN with magnitude spectrum as input for the discriminator to demonstrate the effectiveness of the replacement from it to the mel spectrogram. For calculation, 

\subsection{Results \& Discussions}
\vspace{-.5cm}
\begin{table}[ht]
\caption{\textit{Evaluation Set \RomanNumeralCaps{1}} results. The input is processed by the proposed IP method if it is not specified. }
\label{tab:1}
\centering
\begin{tabular}{llccc}
\hline
\multirow{2}{*}{Model} & \multirow{2}{*}{} & \multicolumn{3}{c}{WER~\%} \\
 &  & Mean & STD & \textless{}=20 \\ \hline
Baseline & \multicolumn{1}{c}{-} & 14.43 & 7.69 & 75.55 \\
\multirow{3}{*}{CMGAN} & Original & 30.34 & 28.49 & 44.86 \\
 & Retrained & 10.29 & 6.06 & 84.99 \\
 & w/o IP & 11.35 & 10.08 &  81.99   \\
\multirow{3}{*}{\begin{tabular}[c]{@{}l@{}}Two-Mask\\ CMGAN\end{tabular}} & proposed & \textbf{7.44} & 3.92 & \textbf{91.40} \\
 & w/o IP & 12.30 & 14.14 & 78.92 \\
 & w/o mel & 9.56 & 5.26 & 86.00 \\ \hline
\end{tabular}
\end{table}

% We further illustrate the result by Fig.~\ref{fig:3}. 
In Table~\ref{tab:1}, we present the results %of the proposed method compared to the baseline model 
of \textit{Evaluation Set \RomanNumeralCaps{1}}. It is observed that the original CMGAN %does not improve the recognition result in distorted speech. Instead, it 
aggravates rather than improves the distortion caused by spectral oversubtraction, increasing the average WER from 14.43\% to 30.34\%. The WERs of less than half of the files are greater than 20\%. Comparing the results of the retrained CMGAN and \textit{w/o mel}, we find that the mask generation strategy of the proposed two-mask performs better than that of the single-mask. The adoption of the mel spectrogram as input to the discriminator improves the performance from 9.56\% to 7.44\%. In addition, if only the audio blocks are directly processed (w/o IP), the performance can be as poor as 12.30\%, while the proposed IP method for streaming blocks can improve the performance by 39.51\%. %It shows that the proposed IP method for streaming audio recording buffers manages to leverage the requirement for real-time processing during HRI and long-term information for comparable performance. Overall, the proposed Two-Mask CMGAN with IP input is more accurate and reliable.

% To compare the performance of the proposed Two-Mask CMGAN and CMGAN after retraining, we selected an estimated human speech after enhancement, shown in Fig~\ref{fig:4}. It can be observed that, in the zone highlighted by the red rectangle, the speech spectrogram enhanced by the proposed network has more details. In fact, the originally misrecognized words, \textit{"hands and knees"}, can be predicted by Whisper in enhanced speech from the proposed network. In contrast, these words are still missing from the transcription of the estimated human interruption speech enhanced by the retrained CMGAN.

% \begin{figure}[ht]
%     \centering
%     \includegraphics[width=0.33\textwidth]{figs/Result_comparison.png}
%     \caption{Spectrogram comparison of the enhanced estimated human speech and the target human speech.}
%     \label{fig:4}
% \end{figure}

% \begin{figure}[ht]
%     \centering
%     \includegraphics[width=0.33\textwidth]{figs/set1.png}
%     \caption{Evaluation result on \textit{Evaluation Set \RomanNumeralCaps{1}}. The value in each box stands for the mean value.}
%     \label{fig:3}
% \end{figure}

\begin{figure}[bth]
    \centering
    \includegraphics[width=0.48\textwidth]{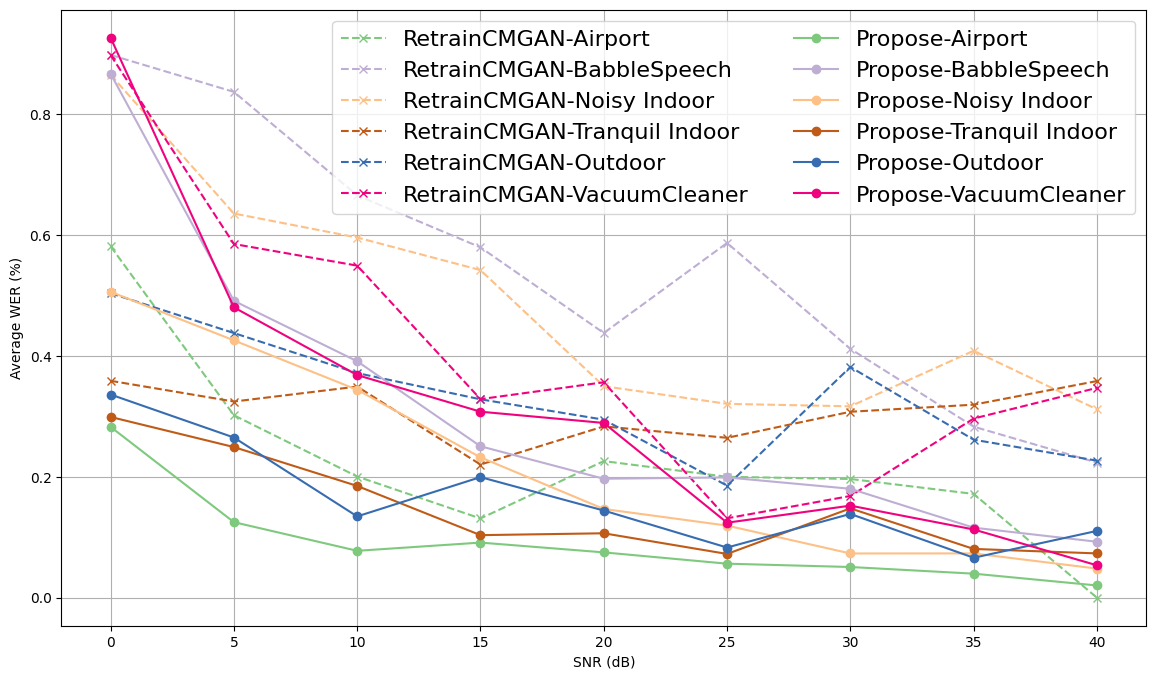}
    \caption{Evaluation result on \textit{Evaluation Set \RomanNumeralCaps{2}}.}
    \label{fig:5}
\end{figure}

The performance on Evaluation set \RomanNumeralCaps{2} for different SNRs is presented in Fig.~\ref{fig:5}. %The WER of the estimated speech interruption is 37\%, that of
% enhanced speech by the proposed network is 4.17\%, and that of the retrained CMGAN is 29.17\%. It is observed that in most scenarios, the performance of the proposed network exceeds that of the retrained CMGAN network. It shows that to enhance the distorted speech suffering from spectral oversubtraction in FFR, learning from the restored values in HFR and generating two masks, one for compensation and one for denoising, is beneficial. %Performing a thorough analysis of the performance in each scenario, 
We find that the proposed method performed particularly well when Airport, Tranquil Indoor and Outdoor groups are mixed in. The average WER is reduced to less than 20\% when the SNR is no less than 10~dB. In more challenging scenarios, such as \textit{Babble Speech} and \textit{Noisy Indoor}, it can still improve the WER to less than 20\% when the SNR is no less than 20~dB. The reason why it performs relatively poorly in these scenarios is that the HRF values of competitive speakers and noises are non-stationary and high. 
The enhancement process will generate relatively more artifacts according to these and will further result in poor recognition. In comparison, the spectrogram of other noise groups has more features in FFR, which is subtracted during the interruption detection by RESF. Especially when there are no competitive speakers or sharp noises present during human interruption, the proposed work can enhance the detected interruption and reduce misrecognized words to a reasonable ratio.

\vspace{-.3cm}

\section{Conclusion}
In this work, we propose a Two-Mask CMGAN network that targets the enhancement of the detected human interruption during robot speech, which suffers from spectral oversubtraction in FFR. We also propose an IP method that allows networks trained on long fixed-length input to process streaming audio blocks. Evaluations of two datasets, including one with unseen noise, demonstrate significant recognition improvements. The combination of Two-Mask CMGAN and IP is verified to improve the robustness of the RESF pipeline.  In conclusion, the combination of RESF and the proposed Two-Mask CMGAN with IP shows potential for deployment 
%Pepper 
in a real-world HRI setting, enabling the robot to keep its single-channel microphone open even when it is speaking. In future work, we will deploy the network in the humanoid robot Pepper and test its effectiveness in such real-world HRI scenarios.

\bibliographystyle{IEEEtran}
\bibliography{mybib}

\begin{thebibliography}{10}
\providecommand{\url}[1]{#1}
\csname url@samestyle\endcsname
\providecommand{\newblock}{\relax}
\providecommand{\bibinfo}[2]{#2}
\providecommand{\BIBentrySTDinterwordspacing}{\spaceskip=0pt\relax}
\providecommand{\BIBentryALTinterwordstretchfactor}{4}
\providecommand{\BIBentryALTinterwordspacing}{\spaceskip=\fontdimen2\font plus
\BIBentryALTinterwordstretchfactor\fontdimen3\font minus \fontdimen4\font\relax}
\providecommand{\BIBforeignlanguage}[2]{{%
\expandafter\ifx\csname l@#1\endcsname\relax
\typeout{** WARNING: IEEEtran.bst: No hyphenation pattern has been}%
\typeout{** loaded for the language `#1'. Using the pattern for}%
\typeout{** the default language instead.}%
\else
\language=\csname l@#1\endcsname
\fi
#2}}
\providecommand{\BIBdecl}{\relax}
\BIBdecl

\bibitem{paliwal2010single}
K.~Paliwal, K.~W{\'o}jcicki, and B.~Schwerin, ``Single-channel speech enhancement using spectral subtraction in the short-time modulation domain,'' \emph{Speech communication}, 2010.

\bibitem{das2021fundamentals}
N.~Das, S.~Chakraborty, J.~Chaki, N.~Padhy, and N.~Dey, ``Fundamentals, present and future perspectives of speech enhancement,'' \emph{International Journal of Speech Technology}, vol.~24, no.~4, pp. 883--901, 2021.

\bibitem{Li_2024}
Y.~Li, K.~Hindriks, and F.~Kunneman, ``Single-channel robot ego-speech filtering during human-robot interaction,'' in \emph{Proceedings of the 2024 International Symposium on Technological Advances in Human-Robot Interaction}, ser. TAHRI 2024.\hskip 1em plus 0.5em minus 0.4em\relax ACM, 2024.

\bibitem{upadhyay2015speech}
N.~Upadhyay and A.~Karmakar, ``Speech enhancement using spectral subtraction-type algorithms: A comparison and simulation study,'' \emph{Procedia Computer Science}, vol.~54, pp. 574--584, 2015.

\bibitem{kleinschmidt2010robust}
T.~F. Kleinschmidt, ``Robust speech recognition using speech enhancement,'' Ph.D. dissertation, Queensland University of Technology, 2010.

\bibitem{li2024near}
Y.~Li, F.~A. Kunneman, and K.~V. Hindriks, ``A near-real-time processing ego speech filtering pipeline designed for speech interruption during human-robot interaction,'' \emph{arXiv preprint arXiv:2405.13477}, 2024.

\bibitem{parikh2005influence}
G.~Parikh and P.~C. Loizou, ``The influence of noise on vowel and consonant cues,'' \emph{The Journal of the Acoustical Society of America}, 2005.

\bibitem{chaudhari2015review}
A.~Chaudhari and S.~Dhonde, ``A review on speech enhancement techniques,'' in \emph{2015 International Conference on Pervasive Computing (ICPC)}.\hskip 1em plus 0.5em minus 0.4em\relax IEEE, 2015, pp. 1--3.

\bibitem{kamath2002multi}
S.~Kamath, P.~Loizou \emph{et~al.}, ``A multi-band spectral subtraction method for enhancing speech corrupted by colored noise.'' in \emph{ICASSP}, vol.~4.\hskip 1em plus 0.5em minus 0.4em\relax Citeseer, 2002, pp. 44\,164--44\,164.

\bibitem{hu2020dccrn}
Y.~Hu, Y.~Liu, S.~Lv, M.~Xing, S.~Zhang, Y.~Fu, J.~Wu, B.~Zhang, and L.~Xie, ``Dccrn: Deep complex convolution recurrent network for phase-aware speech enhancement,'' \emph{arXiv preprint arXiv:2008.00264}, 2020.

\bibitem{hao2021fullsubnet}
X.~Hao, X.~Su, R.~Horaud, and X.~Li, ``Fullsubnet: A full-band and sub-band fusion model for real-time single-channel speech enhancement,'' in \emph{ICASSP 2021-2021 IEEE International Conference on Acoustics, Speech and Signal Processing (ICASSP)}.\hskip 1em plus 0.5em minus 0.4em\relax IEEE, 2021, pp. 6633--6637.

\bibitem{wali2022generative}
A.~Wali, Z.~Alamgir, S.~Karim, A.~Fawaz, M.~B. Ali, M.~Adan, and M.~Mujtaba, ``Generative adversarial networks for speech processing: A review,'' \emph{Computer Speech \& Language}, vol.~72, p. 101308, 2022.

\bibitem{pascual2017segan}
S.~Pascual, A.~Bonafonte, and J.~Serra, ``Segan: Speech enhancement generative adversarial network,'' \emph{arXiv preprint arXiv:1703.09452}, 2017.

\bibitem{fu2019metricgan}
S.-W. Fu, C.-F. Liao, Y.~Tsao, and S.-D. Lin, ``Metricgan: Generative adversarial networks based black-box metric scores optimization for speech enhancement,'' in \emph{International Conference on Machine Learning}.\hskip 1em plus 0.5em minus 0.4em\relax PMLR, 2019, pp. 2031--2041.

\bibitem{cao2022cmgan}
R.~Cao, S.~Abdulatif, and B.~Yang, ``Cmgan: Conformer-based metric gan for speech enhancement,'' \emph{arXiv preprint arXiv:2203.15149}, 2022.

\bibitem{thiemann2013diverse}
J.~Thiemann, N.~Ito, and E.~Vincent, ``The diverse environments multi-channel acoustic noise database (demand): A database of multichannel environmental noise recordings,'' in \emph{Proceedings of Meetings on Acoustics}, vol.~19, no.~1.\hskip 1em plus 0.5em minus 0.4em\relax AIP Publishing, 2013.

\bibitem{ristea2024icassp}
N.~C. Ristea, A.~Saabas, R.~Cutler, B.~Naderi, S.~Braun, and S.~Branets, ``Icassp 2024 speech signal improvement challenge,'' \emph{arXiv preprint arXiv:2401.14444}, 2024.

\bibitem{Donahue2018}
C.~Donahue, B.~Li, and R.~Prabhavalkar, ``Exploring speech enhancement with generative adversarial networks for robust speech recognition,'' in \emph{2018 IEEE International Conference on Acoustics, Speech and Signal Processing (ICASSP)}, 2018, pp. 5024--5028.

\bibitem{titze1998principles}
I.~R. Titze and D.~W. Martin, ``Principles of voice production,'' 1998.

\bibitem{abdulatif2024cmgan}
S.~Abdulatif, R.~Cao, and B.~Yang, ``Cmgan: Conformer-based metric-gan for monaural speech enhancement,'' \emph{IEEE/ACM Transactions on Audio, Speech, and Language Processing}, 2024.

\bibitem{Mao_2017_ICCV}
X.~Mao, Q.~Li, H.~Xie, R.~Y. Lau, Z.~Wang, and S.~Paul~Smolley, ``Least squares generative adversarial networks,'' in \emph{Proceedings of the IEEE International Conference on Computer Vision (ICCV)}, Oct 2017.

\bibitem{bahdanau2014neural}
D.~Bahdanau, K.~Cho, and Y.~Bengio, ``Neural machine translation by jointly learning to align and translate,'' \emph{arXiv preprint arXiv:1409.0473}, 2014.

\bibitem{kim2020t}
J.~Kim, M.~El-Khamy, and J.~Lee, ``T-gsa: Transformer with gaussian-weighted self-attention for speech enhancement,'' in \emph{ICASSP 2020-2020 IEEE International Conference on Acoustics, Speech and Signal Processing (ICASSP)}.\hskip 1em plus 0.5em minus 0.4em\relax IEEE, 2020, pp. 6649--6653.

\bibitem{reddy2019scalable}
C.~K. Reddy, E.~Beyrami, J.~Pool, R.~Cutler, S.~Srinivasan, and J.~Gehrke, ``A scalable noisy speech dataset and online subjective test framework,'' \emph{Proc. Interspeech 2019}, pp. 1816--1820, 2019.

\end{thebibliography}
\end{document}